\documentclass[10pt,twocolumn,letterpaper]{article}

\usepackage{cvpr}              

\makeatletter
\robustify\@latex@warning@no@line
\makeatother

\usepackage{graphicx}
\usepackage{amsmath, bm}
\usepackage{amssymb}
\usepackage{booktabs}
\usepackage{amsthm}
\usepackage{colortbl}
\usepackage{comment}
\usepackage[dvipsnames]{xcolor}
\usepackage[pagebackref,breaklinks,colorlinks]{hyperref}
\usepackage{color}
\usepackage{multirow} 
\usepackage[accsupp]{axessibility}  

\usepackage[capitalize]{cleveref}

\usepackage{colortbl}
\usepackage{hhline}
\usepackage{nicematrix,tikz}
\usepackage{hyperref}

\usepackage{authblk}

\crefname{section}{Sec.}{Secs.}
\Crefname{section}{Section}{Sections}
\Crefname{table}{Table}{Tables}
\crefname{table}{Tab.}{Tabs.}

\title{Focus on Hiders: Exploring Hidden Threats for Enhancing Adversarial Training}
\author[a,b]{Qian Li}
\author[b,d]{Yuxiao Hu}
\author[c]{Yinpeng Dong}
\author[b]{Dongxiao Zhang}
\author[b]{Yuntian Chen\thanks{Corresponding author: ychen@eias.ac.cn}}
\affil[a]{Shanghai Jiao Tong University, Shanghai, China;}
\affil[b]{Eastern Institute of Technology, Ningbo, China;}
\affil[c]{Dept. of Comp. Sci. and Tech., Institute for AI, Tsinghua-Bosch Joint ML Center, THBI Lab, BNRist Center, Tsinghua University, Beijing 100084, China;}
\affil[d]{The Hong Kong Polytechnic University, HongKong, China}

\begin{document}



\maketitle



\newtheorem{theorem}{Theorem}[]
\newtheorem{definition}{Definition}[]
\newtheorem{remark}{Remark}[]

\begin{abstract}
Adversarial training is often formulated as a min-max problem, however, concentrating only on the worst adversarial examples causes alternating repetitive confusion of the model, i.e., previously defended or correctly classified samples are not defensible or accurately classifiable in subsequent adversarial training. We characterize such non-ignorable samples as ``hiders'', which reveal the hidden high-risk regions within the secure area obtained through adversarial training and prevent the model from finding the real worst cases. We demand the model to prevent hiders when defending against adversarial examples for improving accuracy and robustness simultaneously. By rethinking and redefining the min-max optimization problem for adversarial training, we propose a generalized adversarial training algorithm called \textbf{H}ider-\textbf{F}ocused \textbf{A}dversarial \textbf{T}raining (HFAT). HFAT introduces the \textit{iterative evolution optimization strategy} to simplify the optimization problem and employs an auxiliary model to reveal hiders, effectively combining the optimization directions of standard adversarial training and prevention hiders. Furthermore, we introduce an adaptive weighting mechanism that facilitates the model in adaptively adjusting its focus between adversarial examples and hiders during different training periods. We demonstrate the effectiveness of our method based on extensive experiments, and ensure that HFAT can provide higher robustness and accuracy.
\end{abstract}


\section{Introduction}\label{sec:intro}
Although deep neural networks (DNNs) have made significant progress in recent years~\cite{cv,speech,nlp}, they are easily fooled by adversarial examples to make incorrect predictions~\cite{fgsm,pgd,mim,Qian23}. These malicious attacks pose a threat to the security and well-being of individuals, which highlights the importance of adversarial defense efforts. Among them, adversarial training is proven to be the most effective defense method against adversarial attacks~\cite{pgd,jin23}.

\begin{figure}
\centering
\includegraphics[width=0.99\columnwidth]{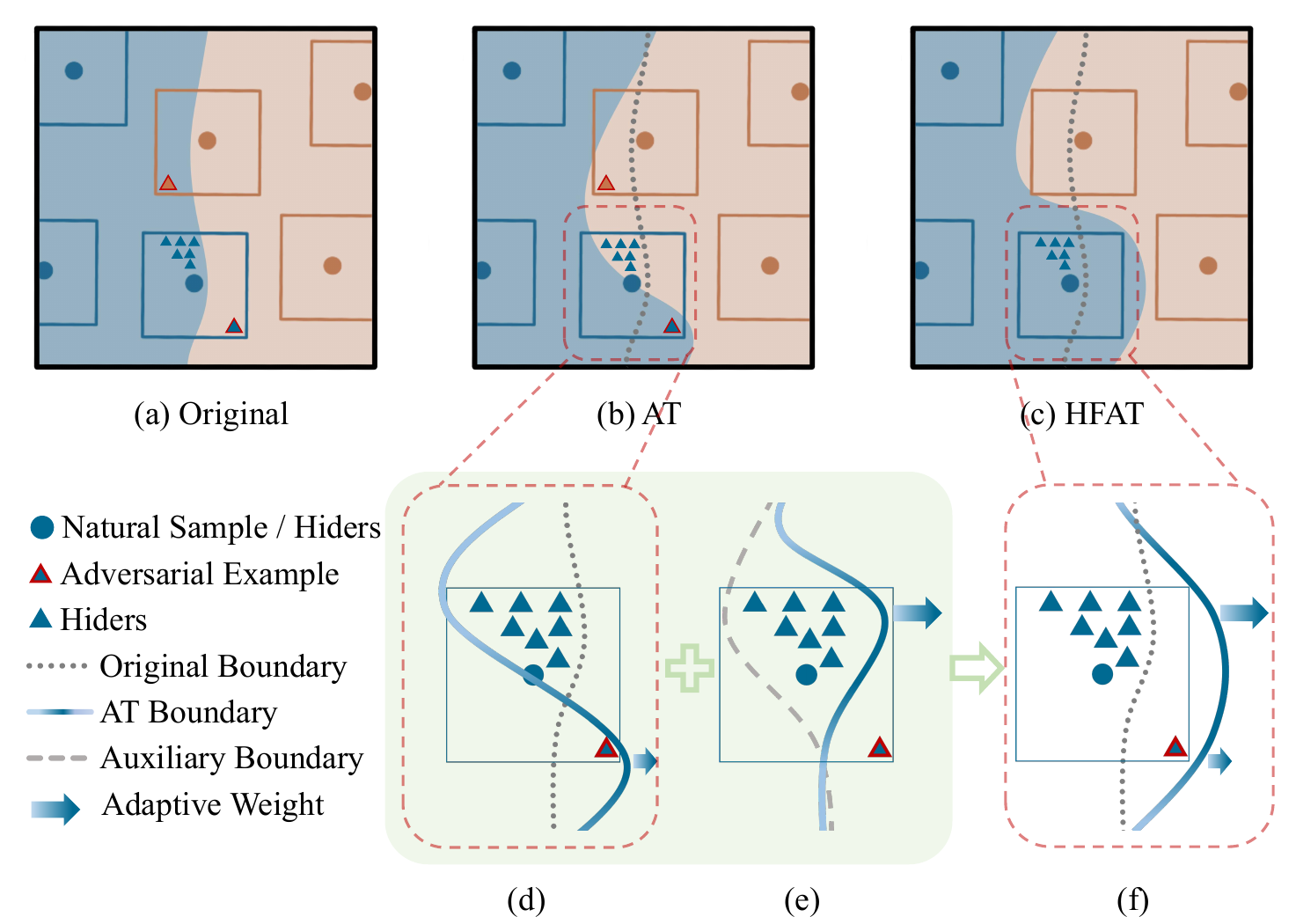}
\caption{Illustration of our core idea. Previous adversarial training methods have single-mindedly concentrated on worst-case adversarial examples, aiming to accurate classification of such examples. However, these methods fail to protect hidden high-risk regions. We refer to these regions' samples as hiders (blue-filled triangles) which are correctly classified in the original model but misclassified after adversarial training due to excessive accommodation of adversarial examples (blue-filled triangles within \textit{original boundary} and outside of \textit{AT boundary}, as depicted in (b), (d)). It is noteworthy to mention that this phenomenon of diminished accuracy also affects natural samples (blue-filled circle), which can be considered as a special type of hiders. By introducing an auxiliary model that exposes the hidden high-risk regions where hiders are located (blue-filled triangles outside of \textit{auxiliary boundary}, as depicted in (e)), we can obtain the optimization direction to prevent hiders. Our core idea is to adaptively defend against both adversarial examples and hiders simultaneously, which promises a defense mechanism that ensures superior robustness and accuracy.}
\vspace{-1.5em}
\label{intro}
\end{figure}

\begin{figure*}
\centering
\includegraphics[width=1\textwidth]{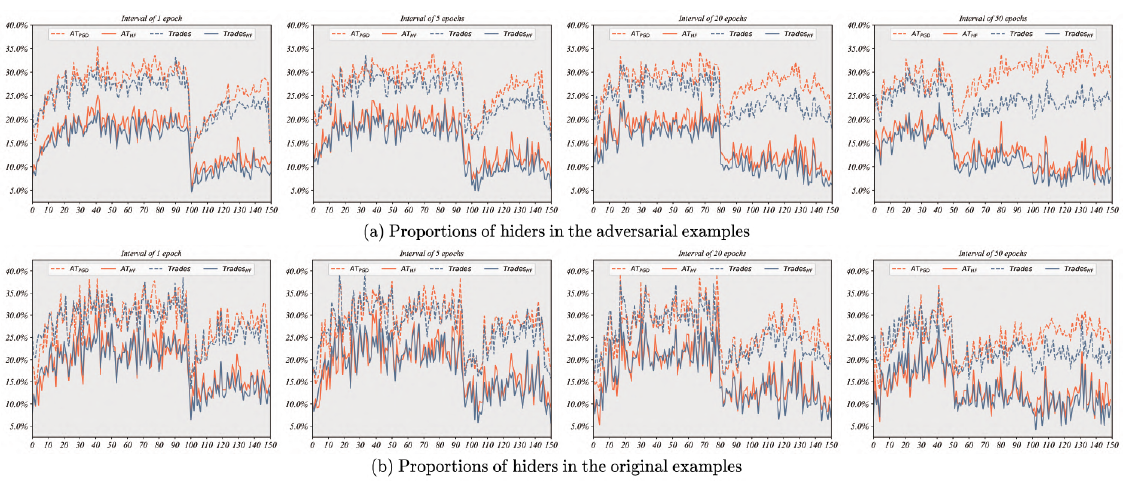}
\caption{Visualization of the proportions of hiders across different epochs. The first row (a) denotes the ratios of hiders in the adversarial examples. The horizontal axis denotes the present epoch. Adversarial examples that are initially defended successfully in the present epoch, but thereafter fail after intervals of 1, 5, 20, and 50 epoch(s), are referred to as hiders. The variations in the proportion of these hiders are depicted in the graphs. We plot the proportions for four different methods: AT$_{\rm PGD}$, AT$_{\rm HF}$, Trades and Trades$_{\rm HF}$ (subscript "HF" represents the proposed approach); The second row (b) follows the same process to depict proportions of hiders in the original samples. For original samples, hiders refer to samples that are first classified correctly in the present epoch but then fail after periods of 1, 5, 20, and 50 epochs. Full details are in Sec.~\ref{ana_pic}.}
\vspace*{-1.5em}
\label{g1}
\end{figure*}

\begin{figure}
\centering
\includegraphics[width=0.49\textwidth]{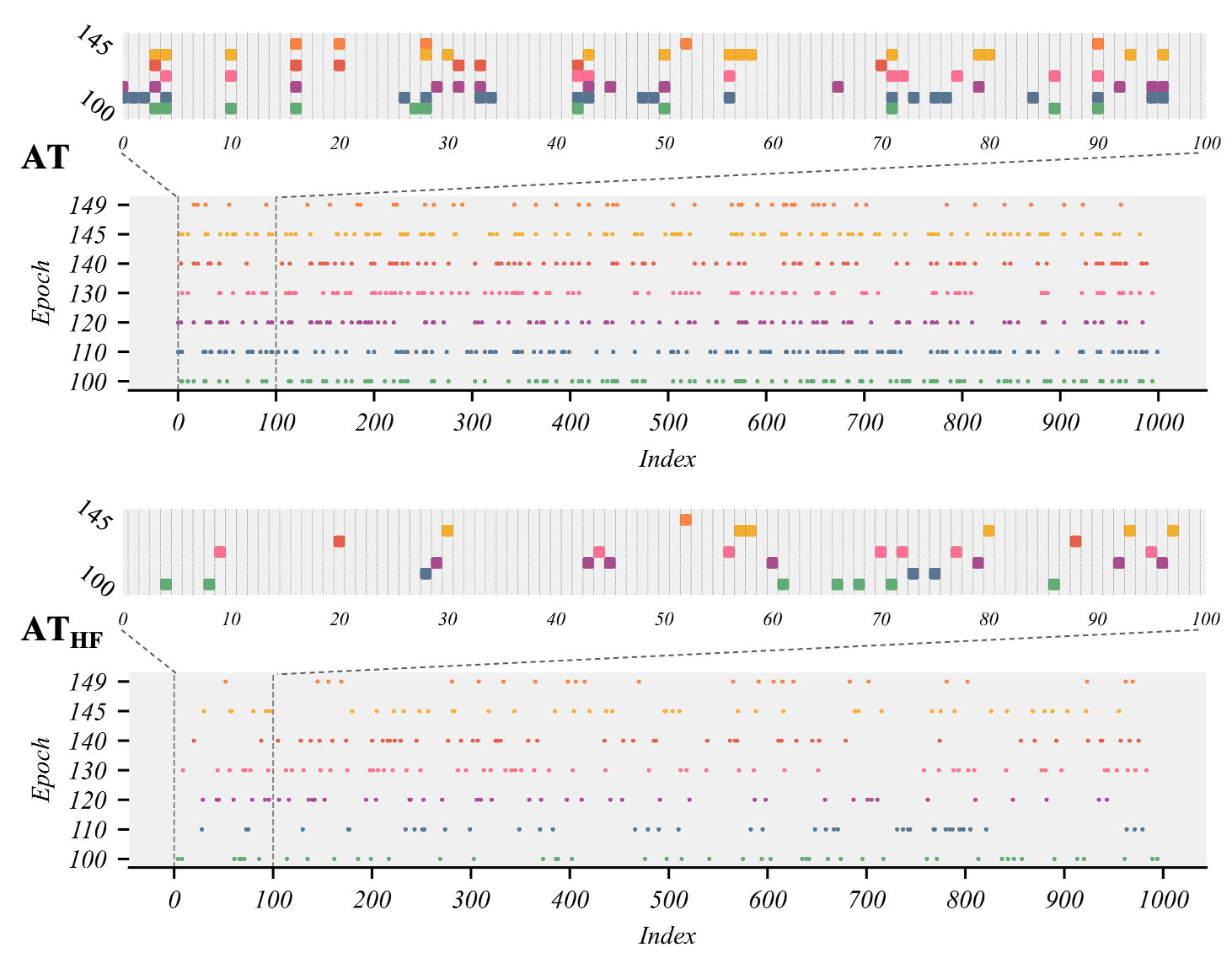}
\caption{Statistical graph of hiders' occurrence locations. A set of 1000 adversarial examples, which the model fails to defend against at epoch 150 using PGD$_{20}$, is collected. If the models from prior epochs (100, 110, 120, 130, 140, 145, and 149) successfully defend these samples, they are labeled as hiders. Additionally, markers are placed at the respective indices. To enhance visualization, we magnify the indices in the range of 0-100 to better discern the disparities between AT$_{\rm PGD}$ and AT$_{\rm HF}$ (AT with our proposed method). Full details are in Sec.~\ref{ana_pic}.}
\vspace{-1.5em}
\label{g2}
\end{figure}

Following the min-max optimization problem~\cite{pgd} of adversarial training, previous methods~\cite{pgd, trades, adt, mart, help, awp} always single-mindedly focus on the worst-case adversarial examples to obtain the optimal solution of inner problem. However, these methods tend to overlook the potential vulnerabilities that also exist in secure areas and result in compromised robustness and accuracy. Concretely, concentrating only on the worst-case adversarial examples causes alternating repetitive confusion of the model as seen in Fig.~\ref{intro}, \ie, adversarial or natural samples that were defended or accurately categorized against in the preceding adversarial training epoch are no longer amenable to defense or accurate classification in the subsequent epoch. Fig.~\ref{g1} demonstrates the ubiquity of this phenomenon throughout all training epochs. Moreover, as the epoch interval increases, a greater number of previously defended samples or accurately classified samples will become susceptible to attacks or misclassification in subsequent epochs. Besides, the observed phenomenon exhibits intermittency. As depicted in Fig.~\ref{g2}, the decision boundary repeatedly confuses certain samples, thereby hindering the model from identifying the genuine worst case (the worst-case adversarial example we find in this epoch may be just a perturbed sample without attack performance for the model in the previous epoch) and impacts the model's performance. We analyze that this phenomenon is caused by single-minded optimization, which excessively adjusts the decision boundary towards adversarial examples and neglects focused protections for the temporary secure regions. This overlooked issue motivates us to reconsider the min-max optimization problem for adversarial training, and explore the possibilities of preventing potential dangers.

In this paper, we first define those non-negligible samples as ``hiders", which were successfully defended or correctly classified in the previous epoch of adversarial training, but exhibit strong attack capability or are misclassified in the later epoch. We propose a new generalized adversarial training algorithm dubbed \textbf{H}iders-\textbf{F}ocused \textbf{A}dversarial \textbf{T}rainning (HFAT), as shown in Fig.~\ref{intro} (c) and (f). HFAT enhances robustness and accuracy by preventing potentially vulnerable areas where hiders are most likely to be identified, while maintaining defense against adversarial examples. Specifically, we redefine the min-max optimization problem and propose the \textit{iterative evolution optimization strategy} to simplify the problem, which allows us to only consider hiders that are relevant to the next epoch. More specifically, by exploring the intrinsic relationship between hiders and adversarial examples, we model the distribution of hiders as a priori knowledge. Given that hiders pose no immediate threat to the current model, directly utilizing them into adversarial training yields limited advantages. Therefore, through probabilistic sampling guided by the prior distribution, we train an auxiliary model that reveals hiders to determine the optimal direction for preventing hiders by adversarial training on auxiliary model. HFAT integrates the adversarial training optimization directions from both the standard model and the auxiliary model to jointly optimize the network. Besides, in order to better couple the two optimization directions during various training phases, we further design an adaptive weighting mechanism that adjusts the emphasis between hiders and adversarial examples in a dynamic manner. In short, the model can boost the optimization weight for an aspect depending on which aspect is currently more needed.

Our contributions are summarized as follows.
\begin{itemize}
  \item [$\bullet$] We first reveal that the single-minded focus of adversarial training on adversarial examples neglects the hidden threats in secure regions that have been successfully defended in the current epoch, resulting in compromised robustness and accuracy.
  \item [$\bullet$] We define hiders and redefine the min-max optimization objective aim to achieve better robustness and accuracy by preventing potentially vulnerable regions while defending against adversarial examples.
  \item [$\bullet$] We propose a generalized adversarial training strategy called Hiders-Focused Adversarial Training (HFAT). HFAT introduces the \textit{iterative evolution optimization strategy} to simplify the optimization problem and employs an auxiliary model to reveal hiders, effectively combining the optimization directions of standard adversarial training and prevention of hiders. Besides, HFAT includes an adaptive weighting mechanism to improve the coupling of the two optimization objectives.
  \item [$\bullet$] We demonstrate the effectiveness of our method based on extensive experiments, and reveal that HFAT effectively mitigates hidden threats posed by hiders.
\end{itemize}

\section{Background and Related work}

\subsection{Adversarial Attack}
The objective of adversarial attacks is to exploit the model's vulnerability in the vicinity of decision boundaries by introducing small, imperceptible perturbations to the inputs, tricking the model into providing incorrect classifications or predictions. Adversarial attack can be represented as the following optimization objective:

$$
\mathop{\rm max}\limits_{{\bm \delta}\in\mathcal{B}(\epsilon)}\mathcal{L}_{\rm CE}(f_{\theta}({\bm x}+{\bm \delta}),{\bm y}),
$$
where $\mathcal{L}_{\rm CE}$ denotes the cross-entropy loss function and $\mathcal{B}(\epsilon)=\{{\bm \delta}: \||{\bm \delta}\||\leq\epsilon\}$ limits the perturbation $\bm{\delta}$ under a certain distance metric (usually $\ell_p$-norm). 

At present, a multitude of adversarial attack methodologies have been proposed, exposing the susceptible components of deep learning models. To increase the efficacy of adversarial perturbations, Projected Gradient Descent (PGD~\cite{pgd}) refines them iteratively. Carlini and Wagner's attack (C\&W~\cite{cw}) formulates an optimization problem with the goal of obtaining misclassification with the fewest possible perturbations. Momentum Iterative Method (MIM~\cite{mim}) enhances traditional iterative optimization techniques by incorporating a momentum term, facilitating more effective and efficient exploration of the adversarial perturbation space. AutoAttack~\cite{aa} presents a suite of diverse attack methods to evaluate model robustness comprehensively. The success of these attack methods makes the adversarial defence a meaningful work for improving model robustness.

\subsection{Adversarial Training}
Adversarial training is an essential approach to enhance the robustness of deep learning models against adversarial attacks. It involves augmenting the training dataset with adversarial examples, forcing the model to learn and defend against adversarial threats. The foundation of adversarial training lies in a min-max optimization framework, which can be formalized as:

$$
\mathop{\rm min}\limits_{\theta}\mathop{\rm max}\limits_{{\bm \delta}\in\mathcal{B}(\epsilon)}\mathcal{L}_{\rm CE}(f_{\theta}({\bm x}+{\bm \delta}),{\bm y}).
$$

Many noteworthy methods have been proposed within the framework of adversarial training. Madry et al. introduced the foundational AT$_{\rm PGD}$~\cite{pgd} framework, focusing on improving robustness of the models. An early stopping variant of AT$_{\rm PGD}$~\cite{es_adv}, proposed by Rice et al., demonstrated notable improvements. Zhang et al. presented the TRADES~\cite{trades} method, exploring a trade-off between standard accuracy and adversarial robustness. Wu et al. delved into the weight loss landscape, introducing Adversarial Weight Perturbation (AWP~\cite{awp}) to effectively enhance model robustness. MART~\cite{mart}, introduced by Wang et al., improved the adversarial example generation process by simultaneously incorporating misclassified clean examples. Jia et al. introduced Learnable Attack Policy Adversarial Training (LAS-AT~\cite{las-at}), a concept that involved learning to automatically generate better attack policies to enhance model robustness.

However, these methodologies still excessively emphasize adversarial examples~\cite{at_weak0, at_weak1}, neglecting the potential threats concealed within secure regions. As a result, these models are repeatedly confused by hiders and unable to identify the genuine worst case, leading to limited robustness and accuracy. In this paper, we enhance the robustness and accuracy of the model by directing our focus towards hidden threats from a new perspective.

\section{Methodology}
\subsection{Hiders}
While adversarial examples directly expose vulnerabilities in the current trained model, hiders reveal hidden threats within the decision boundaries of the model, \ie, samples that were correctly classified or defended in the previous epoch of adversarial training cannot be accurately classified or defended in subsequent epochs. Besides, as illustrated in the second row of Fig.~\ref{g1}, it is noteworthy to emphasize that certain natural samples exhibit hider-like characteristics. Thus, by implementing proactive defense mechanisms against hiders during adversarial training, the accuracy and robustness of the model can be improved simultaneously.

\noindent\textbf{Definitions of hiders.}\quad
We first define the hider $\hat{{\bm x}}={\bm x}+\hat{\bm \delta}^j$ of sample $({\bm x}, {\bm y})$ in the current $i$-th epoch with respect to the later $j$-th epoch as follows: 

$$
D(f_{\theta^i}(\hat{{\bm x}}))=y, D(f_{\theta^j}(\hat{{\bm x}}))\neq y,\quad
i,j\in \{1,2,...\}, j>i,
$$
where $\hat{\bm \delta}^j\in \mathcal{B}(\epsilon)\bigcap S_i$, $S_i$ indicates the interior of the decision boundary in $i$-th epoch, and $\hat{{\bm x}}$ is defended or correctly classified at the $i$-th epoch and fails at the $j$-th epoch. $D$ is the classification function that maps the probability distribution $f_{\theta^i}(\hat{{\bm x}})$ to the class $y$ with the highest probability. 

Similarly to the adversarial examples, for the model of the $i$-th epoch, there exists a worst-case hider $\hat{\bm x}^*$, which can be expressed as
\begin{equation}
\begin{aligned}\label{wch}
\hat{\bm x}^*={\bm x}+\hat{\bm \delta}^*,\quad
(j^*, \hat{\bm \delta}^*)=\mathop{{\rm argmax}}\limits_{j,\hat{\bm \delta}^j}\mathcal{L}_{\rm CE}(f_{\theta^j}({\bm x}+\hat{\bm \delta}^j),{\bm y}).
\end{aligned}
\end{equation}
Unlike the worst-case adversarial example, which indicates the sample with the strongest attack performance under the current model, the worst-case hider indicates the sample within the current decision boundary that exhibits the highest upper bound on its attack performance during the future epochs. As an illustration, let $\hat{\bm x}^*$ represent the worst-case hider of ${\bm x}$ at the $i$-th epoch, indicating that $\hat{\bm x}^*$ lies within the decision boundary under the model $f_{\theta^i}$, and the maximum loss value of $\hat{\bm x}^*$ in the future $j^*$-th epoch is larger than the maximum loss value of any other samples (also within the decision boundary under the model $f_{\theta^i}$) in future epochs. 

\noindent\textbf{Empirical probability distribution of hiders.}\quad
Due to the delayed threat of hiders, it is difficult to pinpoint worst-case hiders. Besides, the distribution of hiders depends on natural samples and models, which encourages us to explore the relative position of hiders from previous adversarial training models rather than absolute location information. We observed a remarkable similarity in the relative positional relationships between hiders and natural/adversarial examples across various adversarial training methods and training phases. This guides us to model the empirical probability distribution~\cite{emp0, emp1, emp2} of hiders' relative position. The distribution not only reveals the position of hiders with natural samples and adversarial examples, but also elucidates the probability that a sample belongs to hiders. 

Based on the observations, we use the Gaussian distribution $G$ to model the relative positional information. Specifically, at the $i$-th epoch, the positional information distribution of hiders relative to the $j$-th epoch is denoted as $G_j$. We will detail the modeling of empirical distributions in Sec.~\ref{distribution}.

\subsection{Hider-Focused Adversarial Training (HFAT)}
To enhance both the robustness and accuracy of the model, our objective is to proactively defend the worst cases involving adversarial examples and hiders. This dual focus can be expressed as the following optimization objective,
\begin{equation}
\begin{aligned}\label{eq1}
{\rm min} [&\mathop{\rm max}\limits_{{\bm \delta}\in\mathcal{B}(\epsilon)}\mathcal{L}_{\rm CE}(f_{\theta^i}({\bm x}+{\bm \delta}),{\bm y})+\\
&\mathop{\rm max}\limits_{j, \hat{\bm\delta}^j\in S^i}\mathcal{L}_{\rm CE}(f_{\theta^j}({\bm x}+\hat{\bm \delta}^j),{\bm y})],
\end{aligned}
\end{equation}
\noindent where the former is optimized for the adversarial examples, the latter for preventing the potential dangers of hiders, and $j$ denotes that the worst-case hider reaches the maximum loss value at the $j$-th epoch. 

The second term of optimization objective \eqref{eq1} forces us to find the maximum loss value of the samples ${\bm x}+\hat{\bm \delta}$ in all future epochs. We propose \textit{Iterative Evolution Optimization Strategy} to simplify the problem. According to Theorem \ref{thm1}, we can optimize objective \eqref{eq1} by considering only the worst-case in the next epoch, \ie, optimization of objective \eqref{eq1} can be simplified into optimizing objective \eqref{eq2}. The proof of Theorem \ref{thm1} is available in the supporting material.
\begin{equation}
\begin{aligned}\label{eq2}
{\rm min} [&\mathop{\rm max}\limits_{{\bm \delta}\in\mathcal{B}(\epsilon)}\mathcal{L}_{\rm CE}(f_{\theta^i}({\bm x}+{\bm \delta}),{\bm y})+\\
&\mathop{\rm max}\limits_{\hat{\bm\delta}^{i+1}\in S^i}\mathcal{L}_{\rm CE}(f_{\theta^{i+1}}({\bm x}+\hat{\bm \delta}^{i+1}),{\bm y})],
\end{aligned}
\end{equation}

\begin{theorem}\label{thm1}
(Iterative Evolution Optimization Strategy) We can optimize objective \eqref{eq1} by iteratively optimizing against the worst-case hider for the next epoch.
\end{theorem}

\noindent\textbf{Auxiliary model.}\quad
The second term of objective \eqref{eq2} serves the purpose of enabling the current model to defend against hidden threatening regions within the current decision boundary. However, optimizing the current model for future scenarios poses a challenge due to the inability to calculate the derivative of the objective \eqref{eq2}'s second term with respect to the current model's weight parameters $\theta^i$. Moreover, if we directly approximate the second term with hiders' loss function $\mathcal{L}_{\rm CE}(f_{\theta^i}({\bm x}+\hat{\bm\delta}^j))$ under model $f_{\theta^i}$, the conventional gradient descent method encounters optimization difficulty since the hiders are not aggressive for model $f_{\theta^i}$.

To address this issue, we obtain an auxiliary model that has a higher loss at ${\bm x}+\hat{\bm\delta}^j$, thereby exposing the region where hiders are located. Furthermore, we enhance the optimization of model $f_{\theta^i}$ by adding the optimization direction from the adversarial training on auxiliary model as momentum. This compels the current model to acquire optimization directions that effectively prevent hiders.

In order to make the auxiliary model $f_{\hat{\theta}^i}$ expose the region where hiders are situated, it is necessary to determine where hiders are most probable to emerge. By sampling from the empirical probability distribution $G$, we can determine the relative location ratio $r$ of hiders between natural samples and adversarial examples. This allows us to find the most probable region where hiders are placed, as sampling is dependent on probability. In particular, since we only need to consider the hiders associated with the next epoch, we sample the relative position ratio $r$ from $G_1$. Then we can obtain the auxiliary model $f_{\hat{\theta}^i}$ through reverse training as follows: 

\vspace{-0.5em}
$$
\hat{\theta}^i\gets\theta^i+\eta\nabla_{\theta^i}\mathcal{L}_{\rm CE}(f_{\theta^i}(T({\bm x},{\bm x}_{\rm adv},r)),{\bm y}),\quad r\sim G_1,
$$
where $\eta$ is the learning rate, $T$ represents a position transformation function that computes the most probable regions for hiders based on sampled $r$, natural samples, and adversarial examples. We incorporate a minor amount of noise within $\epsilon$ into the process to introduce a degree of randomness. Furthermore, by applying adversarial training on auxiliary model, we are able to determine the gradient direction of defending the hidden threatening regions, which can be utilized as an approximation for the second term in objective \eqref{eq2}. We introduce it as momentum $p$ in the training of model $f_{\theta^i}$, where the $p$ can be denoted as:
\begin{equation}
\begin{aligned}\label{pi}
p^i&=\nabla_{\hat{\theta}^i}(\mathcal{L}_{\rm CE}(f_{\hat{\theta}^i}({\bm x}+{\bm\delta^*}),{\bm y})),\\
\quad {\bm\delta^*}&=\mathop{\rm max}\limits_{{\bm\delta^*}\in\mathcal{B}(\epsilon)}\mathcal{L}_{\rm CE}(f_{\hat{\theta}^i}({\bm x}+{\bm\delta^*}),{\bm y}).
\end{aligned}
\end{equation}

HFAT aims to prevent potential threats while defending adversarial examples, so the optimization of HFAT can be expressed as a coupling of standard adversarial training and auxiliary model guided optimization, which can be formalized as:
\begin{equation}
\begin{aligned}\label{opt1}
\theta^{i+1}\gets\theta^i-\eta(\nabla_{\theta^i}\mathcal{L}_{\rm CE}(f_{\theta^i}({\bm x}+{\bm\delta}),y)+p^i).
\end{aligned}
\end{equation}

\noindent\textbf{Adaptive weighting mechanism.}\quad
In fact, HFAT can be conceptualized as consisting of two adversarial training branches. The first branch involves standard adversarial training, where the model is trained to focus on the worst-case adversarial examples. The second branch concentrates on adversarial training the auxiliary model to assist the model in defending against the region with the highest hidden threat. Considering that the threat intensity of adversarial examples and hiders to the model varies across samples and training phases, we devise an adaptive weighting mechanism to improve the coupling between the two adversarial training branches.

We utilize the disparity between the outputs of natural and adversarial examples as a metric, which represents significance of the branch. If there is a significant disparity between the two outputs, it indicates that the branch is comparatively undertrained and requires increased emphasis during training. The Kullback-Leibler divergence is used to quantify the difference. The adaptive weighting mechanism can be expressed as:
\begin{equation}
\begin{aligned}\label{weight}
\lambda_A = \frac{e^{{\rm KL}(f_{\hat{\theta}}({\bm x})||f_{\hat{\theta}}({\bm x'}))}}{e^{{\rm KL}(f_\theta({\bm x})||f_\theta({\bm x'}))}+e^{{\rm KL}(f_{\hat{\theta}}({\bm x})||f_{\hat{\theta}}({\bm x'}))}},
\end{aligned}
\end{equation}
\noindent where $\lambda_A$ denotes the weight of the momentum $p$ from the auxiliary model's branch, and the weight of standard adversarial training's branch is $\lambda_S=1-\lambda_A$. 

\noindent\textbf{Training strategy.}\quad
With the introduction of auxiliary model and adaptive weighting mechanisms, the update of HFAT's weighting parameters finally can be represented as formula \ref{opt2}, where $p^i$ is shown in formula \ref{pi}.
\begin{equation}
\begin{aligned}\label{opt2}
\theta^{i+1}\gets\theta^i-\eta(\lambda_S\nabla_{\theta^i}\mathcal{L}_{\rm CE}(f_{\theta^i}({\bm x}+{\bm\delta}),y)+\lambda_{A}p^i).
\end{aligned}
\end{equation}

\vspace{-0.8em}
\section{Experiments}

\subsection{Experimental setting}
\textbf{Dataset:} We conduct extensive experiments on the CIFAR-10~\cite{cifar}, CIFAR-100~\cite{cifar}, and SVHN~\cite{SVHN} datasets. We employ a perturbation budget of $8 / 255$ for three datasets. \textbf{Network Architectures:} To train on these datasets, we employed a standard network Pre-ResNet18~\cite{preres} and an advanced large-scale network (WideResNet-34-10~\cite{wrn}). \textbf{Baselines:} We adopt a standard defense baseline: AT$_{\rm PGD}$~\cite{pgd} and four strong defense baselines: TRADES~\cite{trades}, MART~\cite{mart}, AWP~\cite{awp}, HELP~\cite{help} to verify generalization in various conditions. \textbf{Training Details}: All defenses undergo 200 epochs of training using SGD with a momentum of 0.9, weight decay of $5 \times 10^{-4}$, and an initial learning rate of 0.1. The learning rate is reduced by a factor of 10 at the 100-th and 150-th epoch. Simple data augmentations, including a $32 \times 32$ random crop with 4-pixel padding and random horizontal flip, are applied during the training process for all methods. 

\begin{figure}
\centering
\includegraphics[width=0.3\textwidth]{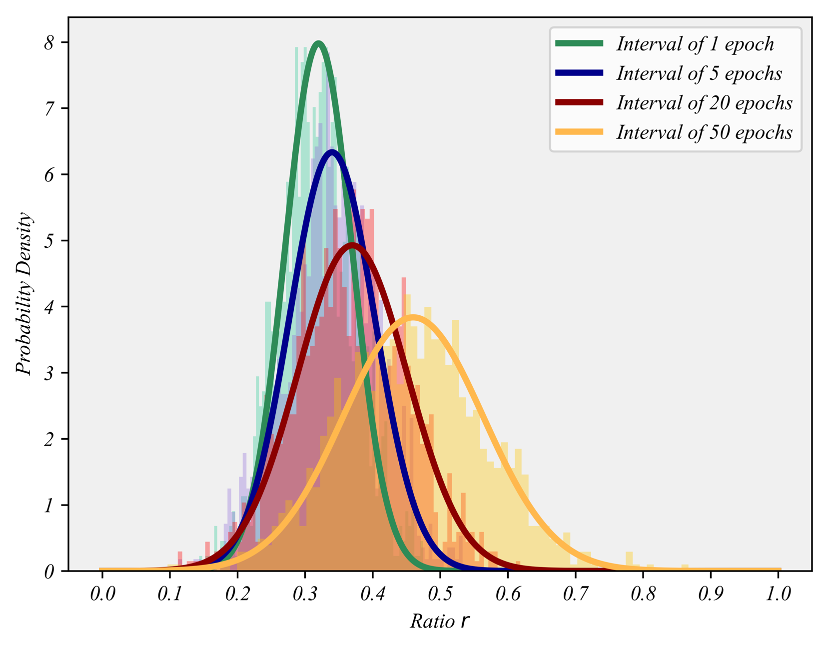}
\caption{Histograms depicting the distribution of 10,000 ratio values ($r$) from five defense models: AT$_{\rm PGD}$, Trades, MART, AWP, and HELP. The ratio $r$ quantifies the distances between hider samples and original samples along the current model's gradient direction, divided by the corresponding distances between adversarial examples and original samples. The various colored curves indicate Gaussian fits that correspond to the histograms at different interval epochs.}
\vspace{-1.5em}
\label{Gauss-dist}
\end{figure}

\subsection{Empirical probability distribution of hiders}\label{distribution}
To obtain the regions where hiders are likely to occur, we compared the distances of both hiders and adversarial examples to the original samples, resulting in the relative position ratio $r$. Fig.~\ref{Gauss-dist} displays the histograms of the ratio $r$ for 10000 hiders, which are computed on five defense models (AT$_{\rm PGD}$~\cite{pgd}, Trades~\cite{trades}, MART~\cite{mart}, AWP~\cite{awp} and HELP~\cite{help}). We fit the data in accordance with the Gaussian distribution characteristics evident in the histogram. Additionally, observation reveals that as the number of epochs increases, the mean and variance of the ratios $r$ for the generated hiders also show an increase. Due to the statistical analysis being conducted on multiple models and a large number of sample points, the distribution characteristics of hiders that we observed exhibit universality.
    
\subsection{Performance analysis}
\subsubsection{Performance on robustness and accuracy}
We utilize two standard attacks, namely FGSM~\cite{fgsm} and PGD~\cite{pgd}, as well as four strong attacks: C\&W~\cite{cw}, MIM~\cite{mim}, $\mathrm{AA}_{\rm rand}$~\cite{aa} (composed of APGD-CE~\cite{aa} and APGD-DLR~\cite{aa}) and $\mathrm{AA}_{\rm standard}$~\cite{aa} (a collection of diverse parameter-free attacks consisting three white-box attacks: APGD-CE~\cite{aa} and APGD-T~\cite{aa} and FAB-T~\cite{fab}, and a black-box attack: Square Attack~\cite{sa}). C\&W uses the margin-based loss function described in \cite{cw} and utilizes PGD for optimization. Specifically, we employ 20 and 100 steps for PGD, 20 steps for MIM, and 30 steps for C\&W. The step size for these attacks is set to $\alpha = \varepsilon / 4$. Tab. \ref{cifar10-bl} showcases the performance enhancement achieved by HFAT across the five defense baselines on CIFAR-10~\cite{cifar}. Results on the CIFAR-100 and SVHN datasets will be presented in the supporting material. We observe that HFAT improves almost both natural accuracy and robust accuracy against various attack methods, confirming the efficiency and applicability of our strategy. 


\begin{table*}
\centering
\caption{Comparison of performance improvement of HFAT applied to five different baselines on the CIFAR-10 dataset and implemented them on the PreAct ResNet-18 and WideResNet34-10 architectures. For testing the attack methods, we selected FGSM, PGD$_{20}$, PGD$_{100}$, CW, MIM, AA$_{\rm rand}$, and AA. Here, AA refers to the standard version of AutoAttack.}
\scriptsize 
\renewcommand{\arraystretch}{1.2} 
\resizebox{\textwidth}{!}{ 
\begin{tabular}{lccccccccccccccccc}
\hline
\multirow{2}{*}{} & \multicolumn{8}{c}{PreAct ResNet-18} &  & \multicolumn{8}{c}{WideResNet34-10} \\ \cline{2-9} \cline{11-18} 
                  & Natural & FGSM & PGD$_{20}$ & PGD$_{100}$ & CW & MIM & AA$_{\rm rand}$ & AA & & Natural & FGSM & PGD$_{20}$ & PGD$_{100}$ & CW & MIM & AA$_{\rm rand}$ & AA \\ \hline
AT$_{\rm PGD}$              & 81.66 & 57.51 & 52.64 & 52.51 & 50.29 & 52.83 & 49.04 & 48.25 & & 86.40 & 61.83 & 55.60 & 55.19 & 54.59 & 55.70 & 52.95 & 52.06 \\
AT$_{\rm HF}$   & \textbf{81.88} & \textbf{60.04} & \textbf{56.39} & \textbf{56.23} & \textbf{52.60} & \textbf{56.46} & \textbf{51.64} & \textbf{50.61} & & \textbf{87.53} & \textbf{65.76} & \textbf{60.06} & \textbf{59.87} & \textbf{57.64} & \textbf{60.18} & \textbf{56.55} & \textbf{55.58} \\ \hline
TRADES           & \textbf{81.24} & 59.24 & 55.71 & 55.37 & 50.45 & 55.63 & 49.95 & 49.20 & & 83.68 & 61.78 & 59.31 & 59.25 & 54.29 & 59.31 & 54.02 & 53.46 \\
TRADES$_{\rm HF}$& 80.39 & \textbf{59.61} & \textbf{57.41} & \textbf{57.12} & \textbf{51.20} & \textbf{56.95} & \textbf{50.96} & \textbf{50.35} & & \textbf{85.38} & \textbf{63.80} & \textbf{61.12} & \textbf{61.03} & \textbf{55.88} & \textbf{61.16} & \textbf{55.62} & \textbf{55.05} \\ \hline
MART             & 80.63 & 59.54 & 56.16 & 55.86 & 50.31 & 55.41 & 50.47 & 49.62 & & 83.98 & 61.32 & 58.43 & 58.12 & 54.74 & 58.07 & 53.64 & 52.36 \\
MART$_{\rm HF}$  & \textbf{81.14} & \textbf{59.73} & \textbf{57.24} & \textbf{56.97} & \textbf{51.11} & \textbf{56.36} & \textbf{51.61} & \textbf{50.74} & &\textbf{84.76} & \textbf{64.03} & \textbf{61.63} & \textbf{61.47} & \textbf{56.18} & \textbf{60.73} & \textbf{55.23} & \textbf{54.77} \\ \hline
AWP              & 80.81 & 59.38 & 55.59 & 55.47 & 51.89 & 55.70 & 51.04 & 50.06 & & 85.65 & 62.75 & 58.82 & 58.69 & 55.56 & 59.24 & 55.39 & 53.61\\
AWP$_{\rm HF}$   & \textbf{81.17} & \textbf{59.83} & \textbf{55.95} & \textbf{55.87} & \textbf{52.31} & \textbf{56.27} & \textbf{51.82} & \textbf{50.28} & & \textbf{86.41} & \textbf{64.18} & \textbf{62.23} & \textbf{62.06} & \textbf{57.42} & \textbf{60.94} & \textbf{56.22} & \textbf{54.95} \\ \hline
HELP             & 80.75 & 59.57 & 56.41 & 56.13 & 52.34 & 56.18 & 50.63 & 49.76 & & 83.69 & 62.63 & 59.48 & 59.11 & 55.82 & 60.02 & 55.40 & 53.98 \\
HELP$_{\rm HF}$  & \textbf{81.27} & \textbf{60.04} & \textbf{57.82} & \textbf{57.50} & \textbf{52.91} & \textbf{57.05} & \textbf{51.24} & \textbf{50.31} & & \textbf{85.21} & \textbf{64.29} & \textbf{62.54} & \textbf{62.21} & \textbf{57.73} & \textbf{61.25} & \textbf{56.71} & \textbf{55.21} \\ \hline
\end{tabular}}
\vspace{-1.5em}
\label{cifar10-bl}
\end{table*}

\subsubsection{Performance under black-box attacks}

\begin{table}
\centering
\caption{Classification accuracy under transfer-based black-box attacks. We use adversarial examples generated by \textit{source model} to attack the \textit{target model}.}
\renewcommand{\arraystretch}{1.2} 
\resizebox{0.5\textwidth}{!}{ 
\begin{NiceTabular}{l*{4}{w{c}{2cm}}}
\hline
\diagbox{Source}{Target} & AT$_{\rm PGD}$ & AT$_{\rm HF}$ & TRADES & TRADES$_{\rm HF}$ \\ \hline
AT$_{\rm PGD}$              & 52.64 & 61.50 & \cellcolor{Yellow!80}62.76 & \cellcolor{Yellow!80}63.61 \\
AT$_{\rm HF}$   & 60.12 & 56.39 & 62.35 & 63.32 \\
TRADES           & \cellcolor{Yellow!80}63.78 & \cellcolor{Yellow!80}64.76 & 55.71 & 63.68 \\
TRADES$_{\rm HF}$& 62.21 & 62.89 & 63.03 & 57.41 \\ \hline

\CodeAfter
  \tikz \draw [very thick, red] (2-|4) -- (2-|5) ;
  \tikz \draw [very thick, red] (4-|4) -- (4-|5) ;
  \tikz \draw [very thick, red] (2-|4) -- (4-|4) ;
  \tikz \draw [very thick, red] (2-|5) -- (4-|5) ;

  \tikz \draw [very thick, red] (4-|2) -- (4-|3) ;
  \tikz \draw [very thick, red] (6-|2) -- (6-|3) ;
  \tikz \draw [very thick, red] (4-|2) -- (6-|2) ;
  \tikz \draw [very thick, red] (4-|3) -- (6-|3) ;

\end{NiceTabular}}
\label{Trans-attack}
\end{table}

Despite the inclusion of black-box attacks in $\mathrm{AA}_{\rm standard}$, we further extend our evaluation to encompass transfer-based black-box attacks~\cite{trans0, trans1, trans2} using PGD-20. We employed adversarial examples generated from the source model to attack the target model. The results in Tab. \ref{Trans-attack} show that the model improved by HFAT achieved better transferability in adversarial attacks (evident in the red box region in the figure, where the HFAT model exhibited a higher success rate when attacking the same target model). Additionally, it demonstrated superior performance in defending against transfer adversarial examples (as depicted by the yellow background region in the figure, where the HFAT model exhibited higher accuracy when faced with adversarial examples from the same source model).

\subsection{Analysis of hiders}
\subsubsection{Defense Performance}
\label{ana_pic}
We illustrate HFAT's defensive performance against hiders through two aspects.

We first verify that HFAT can effectively defend against potential threats from hiders. We compute and plot the proportions of hiders at different intervals (interval 1, 5, 20, and 50 epochs) in Fig.~\ref{g1}. It is observed that our method significantly reduces the proportion of hiders compared to AT$_{\rm PGD}$ and Trades defenses alone. Furthermore, as the interval value increases, the proportion of adversarial hiders for AT$_{\rm PGD}$ and Trades increases, while our method shows almost no difference across different intervals. This indicates that training the model using experience distribution sampling with an interval of 1 epoch is reasonable and ultimately enables better generalization in defending against hiders with larger interval values.

We next verify that HFAT prevents repeated threats to the model by hiders. We visualize the specific index positions at which hiders appear in Fig.~\ref{g2} and observe a phenomenon of repeated occurrences in the AT$_{\rm PGD}$. This repetition appears to follow an intermittent and alternating pattern. However, in the AT$_{\rm HF}$, hiders generally do not exhibit repeated occurrences, indicating that HFAT can effectively and persistently defend against hiders.

\begin{figure}
\centering
\includegraphics[width=0.49\textwidth]{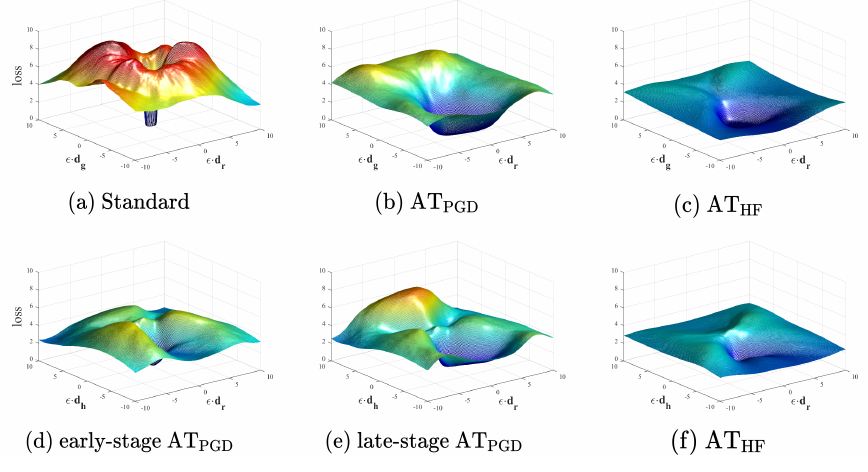}
\caption{The loss surfaces in the vicinity of an input are depicted in (a)-(c) for several models (standard model, AT$_{\rm PGD}$, and AT$_{\rm HF}$).   These entail examining the direction of the gradient (d$_{\rm g}$) and a direction chosen randomly (d$_{\rm r}$)~\cite{adt}. Additionally, in (d)-(f), the loss surfaces focus on the hider direction (d$_{\rm h}$) and a random direction (d$_{\rm r}$), which are depicted for different models (early-stage AT$_{\rm PGD}$ model, late-stage AT$_{\rm PGD}$ model, and AT$_{\rm HF}$).}
\vspace{-2em}
\label{landscape}
\end{figure}

\vspace{-0.8em}
\subsubsection{Loss landscape}
In this subsection, we directly validate the effectiveness of HFAT by visualizing the loss landscape as shown in Fig.~\ref{landscape}~\cite{adt, land0, land1}. Firstly, we perturb the input in the gradient direction and random direction as illustrated in Fig.~\ref{landscape} (a)-(c)~\cite{adt}. It's obvious that HFAT flattens the loss landscape more significantly compared to AT$_{\rm PGD}$, which illustrates that HFAT provides better robustness. Additionally, we visualize the loss landscape along the hider's direction and random direction in Fig.~\ref{landscape} (d)-(f). It can be seen through Fig.~\ref{landscape} (e) that AT$_{\rm PGD}$ exhibits local peaks, which confirms the existence of hiders. However, HFAT effectively suppresses the emergence of hiders as shown in Fig.~\ref{landscape} (f).

\subsection{Ablation studies}

\begin{table}
\centering
\caption{Comparison of robust and natural accuracies under different data augmentation methods. AT$_{\rm hiders}$ includes hider samples directly into the training process as additional training data.}
\scriptsize 
\renewcommand{\arraystretch}{1.2} 
\resizebox{0.45\textwidth}{!}{ 
\begin{tabular}{lcc}
\hline
                    & Robust Accuracy & Natural Accuracy \\ \hline
AT$_{\rm PGD}$                 & 52.64 & 81.66 \\
AT$_{\rm hiders}$  & 48.90 & 85.73 \\
Mixup               & 52.92 & 81.74 \\ \hline
AT$_{\rm HF}$      & 56.39 & 81.88 \\ \hline

\end{tabular}}
\vspace{-1.5em}
\label{da}
\end{table}

\subsubsection{Ablation study of auxiliary model}
Although our method can obtain the regions where hiders are likely to appear by sampling the relative location information, we do not train the model directly using the samples computed by location ratio information as data augmentation, but use the auxiliary model to obtain the optimization direction. Therefore, in this subsection we compare the effects of utilizing auxiliary model and data augmentation. Additionally, we employ the Mixup~\cite{mixup} data augmentation method with a selected $\alpha$ value of 1.4 for further comparison. Tab. \ref{da} reveals that although the direct inclusion of hiders leads to a substantial improvement in natural accuracy, it also results in a significant decrease in robust accuracy. Furthermore, the application of Mixup only yields marginal improvement. Nevertheless, via the implementation of the auxiliary model, HFAT significantly improves both its robustness and accuracy. We believe this is because samples computed by relative location information do not have significant attack performance under the current training epoch, and thus cannot be used directly for training to get good results.

\begin{figure}
\centering
\includegraphics[width=0.48\textwidth]{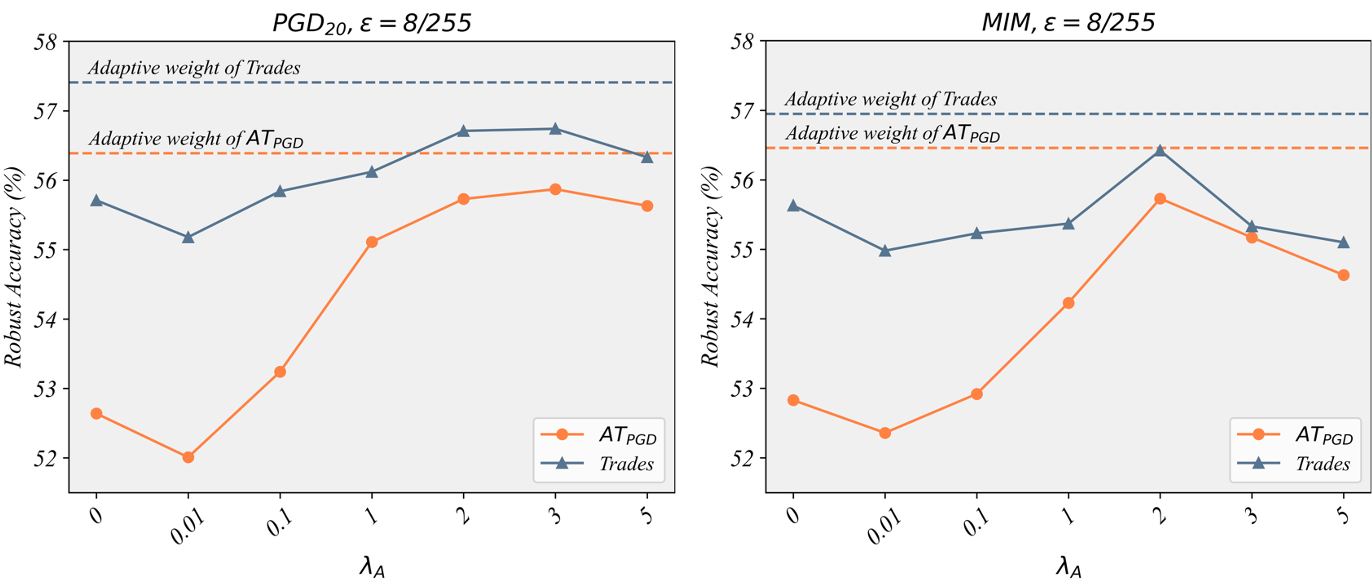}
\caption{The robust accuracy of under white-box attack PGD$_{20}$ and MIM are evaluated using various static auxiliary model weight settings: $\lambda_{A} = 0, 0.01, 0.1, 1, 2, 3$, and 5. The dashed horizontal lines represent the accuracies of AT$_{\rm HF}$ and Trades$_{\rm HF}$ respectively (i.e., using adaptive weighting mechanism). AT$_{\rm PGD}$ and Trades models are represented by distinct curves.}
\vspace{-1em}
\label{weight-ab}
\end{figure}

\vspace{-0.8em}
\subsubsection{Ablation study of adaptive weighting mechanism}
We further investigate the impact of adaptive weighting $\lambda$ on model performance. To convey concepts more effectively, we utilize $\lambda_{A}$ to signify the weight of the auxiliary model branch, and $\lambda_{S}$ to denote the weight of the standard adversarial training branch. In Fig.~\ref{weight-ab}, we set the weight $\lambda_{S}$ to a fixed value of 1, while adjusting $\lambda_{A}$ of the auxiliary model to different values: $\lambda_{A} = 0.0, 0.01, 0.1, 1, 2, 3, 5$. We employ PGD$_{20}$ and MIM attacks to evaluate and represent the robust accuracy of adaptive weights using dashed lines. It is observed that the model's robustness gradually improves as the weight $\lambda_{A}$ increases, reaching its peak when $\lambda_{A}$ equals 2 or 3, followed by a decline. Additionally, the adaptive weight scheme exhibits significantly improved performance compared to static weights. 

\begin{table}
\centering
\caption{Robust accuracy (evaluated by PGD$_{20}$), natural accuracy, and average runtime per epoch under different step value settings. We use AT$_{\rm PGD}$ as a reference.}
\renewcommand{\arraystretch}{1.2} 
\resizebox{0.45\textwidth}{!}{ 
\begin{tabular}{lccc}
\hline
           & Robust Accuracy & Natural Accuracy & Training Time(s)\\ \hline
AT$_{\rm PGD}$        & 52.64 & 81.66 & 192 \\
step 1     & 53.52 & 83.89 & 211 \\
step 3     & 54.26 & 82.75 & 248 \\
step 5     & 56.39 & 81.88 & 287 \\
step 7     & 56.47 & 81.13 & 326 \\ \hline

\end{tabular}}
\vspace{-1.5em}
\label{Step}
\end{table}

\vspace{-0.8em}
\subsubsection{Ablation study of auxiliary model step}
Introducing an auxiliary model incurs additional computational overhead, and to discuss this, we vary the step values for generating adversarial examples of the auxiliary model in Tab. \ref{Step}. Larger step values require more time for computation. As the step value increases, both robust accuracy and natural accuracy gradually improve, reaching their optimal balance when the step value is set to 5. A noticeable decrease in natural accuracy is observed when the step size is set to 7. The reported training time includes the sequential computation for the auxiliary model and the standard adversarial training model. 

\begin{figure}
\centering
\includegraphics[width=0.4\textwidth]{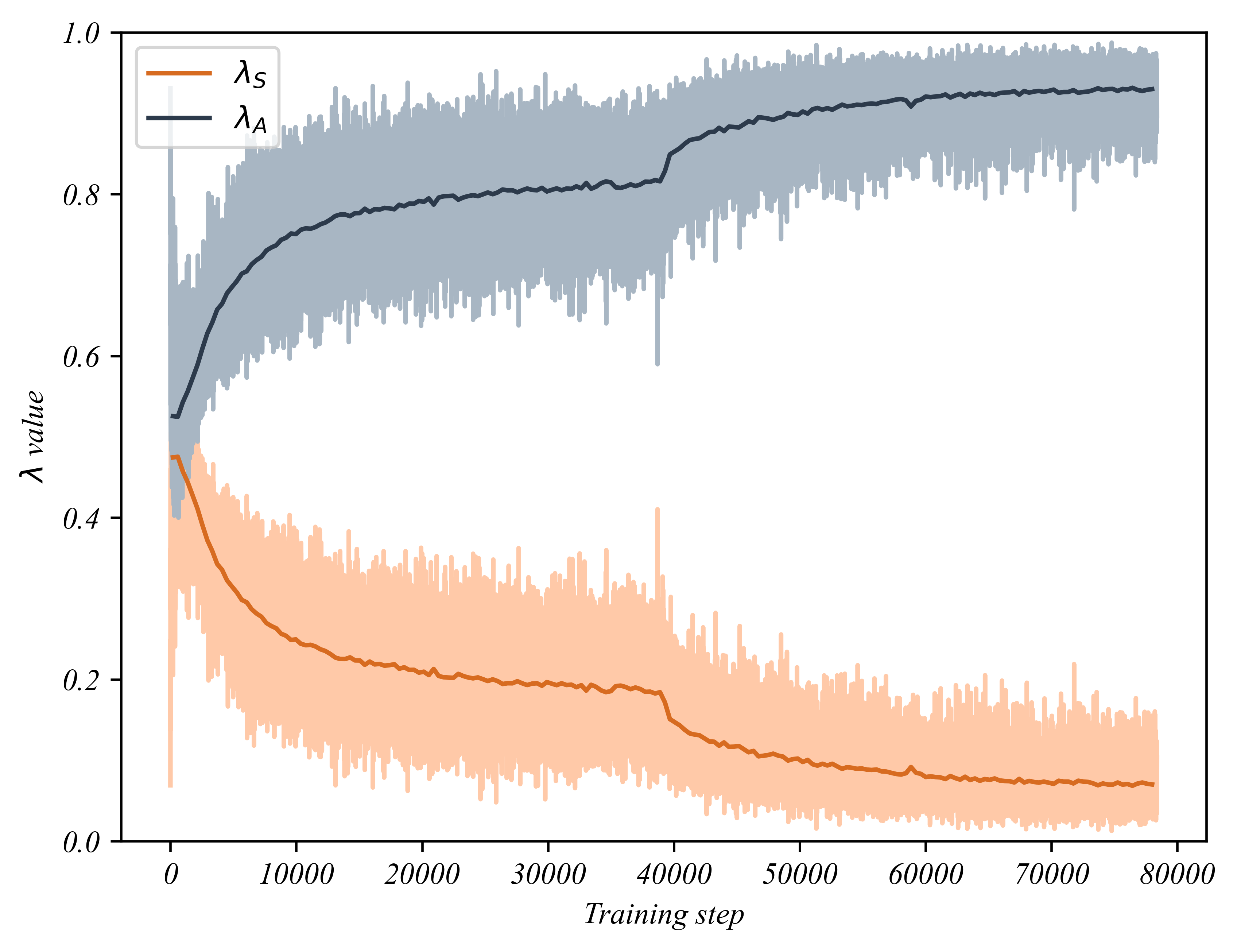}
\vspace{-0.5em}
\caption{Visualization of adaptive weights as they vary with training steps. The blue hue signifies the weights $\lambda_{A}$ of the auxiliary model, whereas the orange hue corresponds to the weights $\lambda_{S}$ of the standard adversarial training branch. In addition, the darker line represents the average value for each epoch.}
\vspace{-1em}
\label{V_aw}
\end{figure}

\subsection{Analysis of adaptive weighting mechanism}
To provide evidence for the effectiveness of introducing an auxiliary model, we visualize the variations of adaptive weight values throughout the training process. Fig.~\ref{V_aw} illustrates the evolution of the mean weight, $\lambda_{S}$, of the standard adversarial training model from around 0.5 in the early epochs to approximately 0.1 in the later stages of training. Conversely, the gradient weight $\lambda_{A}$ of the auxiliary model gradually increases from 0.5 to 0.9. This observation indicates that as the training progresses, the gradient contribution of the auxiliary model becomes increasingly significant, highlighting the heightened importance of hiders' defense in the later stages of model training.

\section{Future work}
We identify hiders as hidden high-risk regions and propose HFAT that defends against both hiders and adversarial examples through iterative evolution optimization strategy and auxiliary model. In this paper, hiders identification relies on empirical fitting based on Gaussian distribution. Thus, we encourage introducing an evaluation metric to assess the model's capability in detecting hidden threats, which could advance research on self-supervision-based methods for hider prevention. Regarding training consumption, since the computations for the auxiliary model and the standard adversarial training model are independent of each other, parallelization can be employed to improve the computational efficiency.

\section{Conclusion}
In this paper, we highlight the limitations of conventional adversarial training, which focuses solely on worst-case adversarial examples and neglects the hidden threats in secure regions. We introduce "hiders" samples that can be defended or correctly classified in earlier epochs but are vulnerable later. Our proposed strategy, HFAT incorporates an auxiliary model to unveil potential threats and offer optimized guidance for enhancing robustness in the regions prone to hider emergence. HFAT enhances model robustness and accuracy by jointly optimizing directions from both models, supported by an adaptive weighting mechanism. Our experiments not only show that HFAT can provide stronger robustness and better accuracy, but also demonstrate HFAT's effectiveness in mitigating hider-related risks.

\clearpage

{
    \small
    \bibliographystyle{ieeenat_fullname}
    \bibliography{main}

\begin{thebibliography}{34}
\providecommand{\natexlab}[1]{#1}
\providecommand{\url}[1]{\texttt{#1}}
\expandafter\ifx\csname urlstyle\endcsname\relax
  \providecommand{\doi}[1]{doi: #1}\else
  \providecommand{\doi}{doi: \begingroup \urlstyle{rm}\Url}\fi

\bibitem[Andriushchenko et~al.(2020)Andriushchenko, Croce, Flammarion, and Hein]{sa}
Maksym Andriushchenko, Francesco Croce, Nicolas Flammarion, and Matthias Hein.
\newblock Square attack: a query-efficient black-box adversarial attack via random search.
\newblock In \emph{European conference on computer vision}, pages 484--501. Springer, 2020.

\bibitem[Bai et~al.(2021)Bai, Luo, Zhao, Wen, and Wang]{at_weak1}
Tao Bai, Jinqi Luo, Jun Zhao, Bihan Wen, and Qian Wang.
\newblock Recent advances in adversarial training for adversarial robustness.
\newblock \emph{arXiv preprint arXiv:2102.01356}, 2021.

\bibitem[Carlini and Wagner(2017)]{cw}
Nicholas Carlini and David Wagner.
\newblock Towards evaluating the robustness of neural networks.
\newblock In \emph{2017 ieee symposium on security and privacy (sp)}, pages 39--57. Ieee, 2017.

\bibitem[Croce and Hein(2020{\natexlab{a}})]{aa}
Francesco Croce and Matthias Hein.
\newblock Reliable evaluation of adversarial robustness with an ensemble of diverse parameter-free attacks.
\newblock In \emph{International conference on machine learning}, pages 2206--2216. PMLR, 2020{\natexlab{a}}.

\bibitem[Croce and Hein(2020{\natexlab{b}})]{fab}
Francesco Croce and Matthias Hein.
\newblock Minimally distorted adversarial examples with a fast adaptive boundary attack.
\newblock In \emph{International Conference on Machine Learning}, pages 2196--2205. PMLR, 2020{\natexlab{b}}.

\bibitem[Devlin et~al.(2018)Devlin, Chang, Lee, and Toutanova]{nlp}
Jacob Devlin, Ming-Wei Chang, Kenton Lee, and Kristina Toutanova.
\newblock Bert: Pre-training of deep bidirectional transformers for language understanding.
\newblock \emph{arXiv preprint arXiv:1810.04805}, 2018.

\bibitem[Dong et~al.(2018)Dong, Liao, Pang, Su, Zhu, Hu, and Li]{mim}
Yinpeng Dong, Fangzhou Liao, Tianyu Pang, Hang Su, Jun Zhu, Xiaolin Hu, and Jianguo Li.
\newblock Boosting adversarial attacks with momentum.
\newblock In \emph{Proceedings of the IEEE conference on computer vision and pattern recognition}, pages 9185--9193, 2018.

\bibitem[Dong et~al.(2020)Dong, Deng, Pang, Zhu, and Su]{adt}
Yinpeng Dong, Zhijie Deng, Tianyu Pang, Jun Zhu, and Hang Su.
\newblock Adversarial distributional training for robust deep learning.
\newblock \emph{Advances in Neural Information Processing Systems}, 33:\penalty0 8270--8283, 2020.

\bibitem[Goodfellow et~al.(2014)Goodfellow, Shlens, and Szegedy]{fgsm}
Ian~J Goodfellow, Jonathon Shlens, and Christian Szegedy.
\newblock Explaining and harnessing adversarial examples.
\newblock \emph{arXiv preprint arXiv:1412.6572}, 2014.

\bibitem[Gowal et~al.(2020)Gowal, Qin, Uesato, Mann, and Kohli]{land1}
Sven Gowal, Chongli Qin, Jonathan Uesato, Timothy Mann, and Pushmeet Kohli.
\newblock Uncovering the limits of adversarial training against norm-bounded adversarial examples.
\newblock \emph{arXiv preprint arXiv:2010.03593}, 2020.

\bibitem[He et~al.(2016{\natexlab{a}})He, Zhang, Ren, and Sun]{cv}
Kaiming He, Xiangyu Zhang, Shaoqing Ren, and Jian Sun.
\newblock Deep residual learning for image recognition.
\newblock In \emph{Proceedings of the IEEE conference on computer vision and pattern recognition}, pages 770--778, 2016{\natexlab{a}}.

\bibitem[He et~al.(2016{\natexlab{b}})He, Zhang, Ren, and Sun]{preres}
Kaiming He, Xiangyu Zhang, Shaoqing Ren, and Jian Sun.
\newblock Identity mappings in deep residual networks.
\newblock In \emph{Computer Vision--ECCV 2016: 14th European Conference, Amsterdam, The Netherlands, October 11--14, 2016, Proceedings, Part IV 14}, pages 630--645. Springer, 2016{\natexlab{b}}.

\bibitem[Jia et~al.(2022)Jia, Zhang, Wu, Ma, Wang, and Cao]{las-at}
Xiaojun Jia, Yong Zhang, Baoyuan Wu, Ke Ma, Jue Wang, and Xiaochun Cao.
\newblock Las-at: adversarial training with learnable attack strategy.
\newblock In \emph{Proceedings of the IEEE/CVF Conference on Computer Vision and Pattern Recognition}, pages 13398--13408, 2022.

\bibitem[Jin et~al.(2023)Jin, Yi, Wu, Mu, and Huang]{jin23}
Gaojie Jin, Xinping Yi, Dengyu Wu, Ronghui Mu, and Xiaowei Huang.
\newblock Randomized adversarial training via taylor expansion.
\newblock In \emph{Proceedings of the IEEE/CVF Conference on Computer Vision and Pattern Recognition}, pages 16447--16457, 2023.

\bibitem[Krizhevsky et~al.(2009)Krizhevsky, Hinton, et~al.]{cifar}
Alex Krizhevsky, Geoffrey Hinton, et~al.
\newblock Learning multiple layers of features from tiny images.
\newblock 2009.

\bibitem[Li et~al.(2023)Li, Hu, Liu, Zhang, Jin, and Chen]{Qian23}
Qian Li, Yuxiao Hu, Ye Liu, Dongxiao Zhang, Xin Jin, and Yuntian Chen.
\newblock Discrete point-wise attack is not enough: Generalized manifold adversarial attack for face recognition.
\newblock In \emph{Proceedings of the IEEE/CVF Conference on Computer Vision and Pattern Recognition}, pages 20575--20584, 2023.

\bibitem[Liu et~al.(2020)Liu, Salzmann, Lin, Tomioka, and S{\"u}sstrunk]{land0}
Chen Liu, Mathieu Salzmann, Tao Lin, Ryota Tomioka, and Sabine S{\"u}sstrunk.
\newblock On the loss landscape of adversarial training: Identifying challenges and how to overcome them.
\newblock \emph{Advances in Neural Information Processing Systems}, 33:\penalty0 21476--21487, 2020.

\bibitem[Madry et~al.(2017)Madry, Makelov, Schmidt, Tsipras, and Vladu]{pgd}
Aleksander Madry, Aleksandar Makelov, Ludwig Schmidt, Dimitris Tsipras, and Adrian Vladu.
\newblock Towards deep learning models resistant to adversarial attacks.
\newblock \emph{arXiv preprint arXiv:1706.06083}, 2017.

\bibitem[Masson and Denoeux(2006)]{emp0}
Marie-H{\'e}l{\`e}ne Masson and Thierry Denoeux.
\newblock Inferring a possibility distribution from empirical data.
\newblock \emph{Fuzzy sets and systems}, 157\penalty0 (3):\penalty0 319--340, 2006.

\bibitem[Netzer et~al.(2011)Netzer, Wang, Coates, Bissacco, Wu, and Ng]{SVHN}
Yuval Netzer, Tao Wang, Adam Coates, Alessandro Bissacco, Bo Wu, and Andrew~Y Ng.
\newblock Reading digits in natural images with unsupervised feature learning.
\newblock 2011.

\bibitem[Rade and Moosavi-Dezfooli(2021)]{help}
Rahul Rade and Seyed-Mohsen Moosavi-Dezfooli.
\newblock Reducing excessive margin to achieve a better accuracy vs. robustness trade-off.
\newblock In \emph{International Conference on Learning Representations}, 2021.

\bibitem[Raghunathan et~al.(2019)Raghunathan, Xie, Yang, Duchi, and Liang]{at_weak0}
Aditi Raghunathan, Sang~Michael Xie, Fanny Yang, John~C Duchi, and Percy Liang.
\newblock Adversarial training can hurt generalization.
\newblock \emph{arXiv preprint arXiv:1906.06032}, 2019.

\bibitem[Rice et~al.(2020)Rice, Wong, and Kolter]{es_adv}
Leslie Rice, Eric Wong, and Zico Kolter.
\newblock Overfitting in adversarially robust deep learning.
\newblock In \emph{International Conference on Machine Learning}, pages 8093--8104. PMLR, 2020.

\bibitem[Saito et~al.(2002)Saito, Coifman, Geshwind, and Warner]{emp1}
Naoki Saito, Ronald~R Coifman, Frank~B Geshwind, and Fred Warner.
\newblock Discriminant feature extraction using empirical probability density estimation and a local basis library.
\newblock \emph{Pattern Recognition}, 35\penalty0 (12):\penalty0 2841--2852, 2002.

\bibitem[Salman et~al.(2020)Salman, Ilyas, Engstrom, Kapoor, and Madry]{trans0}
Hadi Salman, Andrew Ilyas, Logan Engstrom, Ashish Kapoor, and Aleksander Madry.
\newblock Do adversarially robust imagenet models transfer better?
\newblock \emph{Advances in Neural Information Processing Systems}, 33:\penalty0 3533--3545, 2020.

\bibitem[Spanos(2019)]{emp2}
Aris Spanos.
\newblock \emph{Probability theory and statistical inference: Empirical modeling with observational data}.
\newblock Cambridge University Press, 2019.

\bibitem[Tram{\`e}r et~al.(2017)Tram{\`e}r, Kurakin, Papernot, Goodfellow, Boneh, and McDaniel]{trans2}
Florian Tram{\`e}r, Alexey Kurakin, Nicolas Papernot, Ian Goodfellow, Dan Boneh, and Patrick McDaniel.
\newblock Ensemble adversarial training: Attacks and defenses.
\newblock \emph{arXiv preprint arXiv:1705.07204}, 2017.

\bibitem[Wang et~al.(2017)Wang, Deng, Pu, and Huang]{speech}
Yisen Wang, Xuejiao Deng, Songbai Pu, and Zhiheng Huang.
\newblock Residual convolutional ctc networks for automatic speech recognition.
\newblock \emph{arXiv preprint arXiv:1702.07793}, 2017.

\bibitem[Wang et~al.(2019)Wang, Zou, Yi, Bailey, Ma, and Gu]{mart}
Yisen Wang, Difan Zou, Jinfeng Yi, James Bailey, Xingjun Ma, and Quanquan Gu.
\newblock Improving adversarial robustness requires revisiting misclassified examples.
\newblock In \emph{International conference on learning representations}, 2019.

\bibitem[Wu et~al.(2020)Wu, Xia, and Wang]{awp}
Dongxian Wu, Shu-Tao Xia, and Yisen Wang.
\newblock Adversarial weight perturbation helps robust generalization.
\newblock \emph{Advances in Neural Information Processing Systems}, 33:\penalty0 2958--2969, 2020.

\bibitem[Xie et~al.(2019)Xie, Zhang, Zhou, Bai, Wang, Ren, and Yuille]{trans1}
Cihang Xie, Zhishuai Zhang, Yuyin Zhou, Song Bai, Jianyu Wang, Zhou Ren, and Alan~L Yuille.
\newblock Improving transferability of adversarial examples with input diversity.
\newblock In \emph{Proceedings of the IEEE/CVF conference on computer vision and pattern recognition}, pages 2730--2739, 2019.

\bibitem[Zagoruyko and Komodakis(2016)]{wrn}
Sergey Zagoruyko and Nikos Komodakis.
\newblock Wide residual networks.
\newblock \emph{arXiv preprint arXiv:1605.07146}, 2016.

\bibitem[Zhang et~al.(2017)Zhang, Cisse, Dauphin, and Lopez-Paz]{mixup}
Hongyi Zhang, Moustapha Cisse, Yann~N Dauphin, and David Lopez-Paz.
\newblock mixup: Beyond empirical risk minimization.
\newblock \emph{arXiv preprint arXiv:1710.09412}, 2017.

\bibitem[Zhang et~al.(2019)Zhang, Yu, Jiao, Xing, El~Ghaoui, and Jordan]{trades}
Hongyang Zhang, Yaodong Yu, Jiantao Jiao, Eric Xing, Laurent El~Ghaoui, and Michael Jordan.
\newblock Theoretically principled trade-off between robustness and accuracy.
\newblock In \emph{International conference on machine learning}, pages 7472--7482. PMLR, 2019.

\end{thebibliography}
}


\clearpage
\onecolumn

\section{Proof of Theorem 1}
\begin{proof}
    Let $\hat{\bm x}^*={\bm x}+\hat{\bm \delta}^*$ denotes the worst-case hider of sample $({\bm x},{\bm y})$ at the $i$-th epoch, $\hat{\bm \delta}^*\in\mathcal{B}(\epsilon)\cap S_i$ and $\hat{\bm x}^*$ has the highest loss value at the $j$-th epoch. $S_i$ denotes the perturbation set that every $\hat{\bm\delta}\in S_i$ make $\hat{\bm x}={\bm x}+\hat{\bm\delta}$ within the decision boundary at the $i$-th epoch.

    (1) If $j=i+1$, then the theorem follows naturally.

    (2) If $j>i+1$, then there must exist an epoch $l$, $i+1\leq l< j$, satisfy that $\hat{\bm \delta}^*\in S_l$ at the $l$-th epoch, and $\hat{\bm \delta}^*\notin S_{l+1}$ at the $(l+1)$-th epoch, \ie, $\hat{\bm x}^*$ is not an adversarial example at the $l$-th epoch, but becomes an adversarial example at the $(l+1)$-th epoch. If $\hat{\bm x}^*$ is the worst-case hider at the $l$-th epoch, then we can optimize the $\hat{\bm x}^*$ at the $l$-th epoch. If $\hat{\bm x}^*$ is not the worst-case hider at the $l$-th epoch, which suggests that $\hat{\bm x}^*$ no longer has the highest upper bound on its attack performance during the future epochs, then we can indicate $\hat{\bm x}^*$ has been indirectly defended between the $i$-th and $l$-th epochs.

\end{proof}

\section{Performance on robustness and accuracy of CIFAR-100 and SVHN}
\begin{table*}[!ht]
\centering
\caption{Comparison of performance improvement of HFAT applied to five different baselines on the CIFAR-100 and implemented them on the PreAct ResNet-18 and WideResNet34-10 architectures.}
\scriptsize 
\renewcommand{\arraystretch}{1.2} 
\resizebox{\textwidth}{!}{ 
\begin{tabular}{lccccccccccccccccc}
\hline
\multirow{2}{*}{} & \multicolumn{8}{c}{PreAct ResNet-18} &  & \multicolumn{8}{c}{WideResNet34-10} \\ \cline{2-9} \cline{11-18} 
                  & Natural & FGSM & PGD$_{20}$ & PGD$_{100}$ & CW & MIM & AA$_{\rm rand}$ & AA & & Natural & FGSM & PGD$_{20}$ & PGD$_{100}$ & CW & MIM & AA$_{\rm rand}$ & AA \\ \hline
AT$_{\rm PGD}$  & 55.57 & 31.57 & 28.79 & 28.66 & 26.77 & 28.91 & 25.71 & 24.44 & & 58.81 & 33.75 & 30.78 & 30.63 & 29.41 & 30.80 & 27.11 & 25.84 \\
AT$_{\rm HF}$     & \textbf{57.45} & \textbf{34.91} & \textbf{32.32} & \textbf{32.31} & \textbf{29.43} & \textbf{32.38} & \textbf{28.51} & \textbf{27.15} & & \textbf{58.99} & \textbf{37.07} & \textbf{34.47} & \textbf{34.43} & \textbf{31.18} & \textbf{34.57} & \textbf{30.24} & \textbf{28.65} \\ \hline
TRADES          & 54.61 & 32.86 & 30.10 & 29.97 & 27.24 & 30.18 & 26.18 & 25.81 & & 55.54 & 34.77 & 32.86 & 32.81 & 31.37 & 31.88 & 28.57 & 27.36 \\
TRADES$_{\rm HF}$ & \textbf{55.75} & \textbf{34.96} & \textbf{32.29} & \textbf{32.19} & \textbf{29.06} & \textbf{31.41} & \textbf{27.67} & \textbf{27.00} & & \textbf{58.70} & \textbf{35.59} & \textbf{34.49} & \textbf{34.45} & \textbf{32.63} & \textbf{32.49} & \textbf{31.61} & \textbf{30.29} \\ \hline
MART            & 53.41 & 32.85 & 30.53 & 30.39 & 27.93 & 29.77 & 26.12 & 25.31 & & 53.84 & 34.42 & 31.62 & 31.51 & 30.14 & 31.64 & 28.27 & 27.08 \\
MART$_{\rm HF}$   & \textbf{54.74} & \textbf{35.02} & \textbf{32.78} & \textbf{32.55} & \textbf{29.82} & \textbf{30.62} & \textbf{27.93} & \textbf{27.51} & & \textbf{56.19} & \textbf{35.31} & \textbf{33.52} & \textbf{33.39} & \textbf{31.40} & \textbf{32.22} & \textbf{31.49} & \textbf{30.34} \\ \hline
AWP             & 54.19 & 33.21 & 30.71 & 30.62 & 28.05 & 29.93 & 26.54 & 25.49 & & 54.58 & 34.03 & 32.24 & 32.04 & 30.67 & 31.53 & 29.63 & 28.75 \\
AWP$_{\rm HF}$    & \textbf{55.37} & \textbf{35.21} & \textbf{33.14} & \textbf{33.01} & \textbf{30.16} & \textbf{31.56} & \textbf{27.61} & \textbf{27.85} & & \textbf{57.16} & \textbf{35.11} & \textbf{33.02} & \textbf{32.94} & \textbf{31.09} & \textbf{33.06} & \textbf{31.87} & \textbf{31.20} \\ \hline
HELP            & 54.17 & 33.56 & 31.15 & 30.96 & 28.42 & 29.86 & 26.17 & 25.61 & & 55.32 & 34.47 & 31.54 & 31.51 & 29.75 & 31.65 & 29.33 & 28.19 \\
HELP$_{\rm HF}$   & \textbf{55.23} & \textbf{35.46} & \textbf{33.65} & \textbf{33.32} & \textbf{30.24} & \textbf{32.07} & \textbf{28.26} & \textbf{27.87} & & \textbf{58.05} & \textbf{36.40} & \textbf{34.42} & \textbf{34.38} & \textbf{32.58} & \textbf{33.27} & \textbf{32.58} & \textbf{31.36} \\ \hline
\end{tabular}}
\label{cifar100-bl}
\end{table*}

\begin{table*}[!ht]
\centering
\caption{Comparison of performance improvement of HFAT applied to five different baselines on the SVHN and implemented them on the PreAct ResNet-18 and WideResNet34-10 architectures.}
\scriptsize 
\renewcommand{\arraystretch}{1.2} 
\resizebox{\textwidth}{!}{ 
\begin{tabular}{lccccccccccccccccc}
\hline
\multirow{2}{*}{} & \multicolumn{8}{c}{PreAct ResNet-18} &  & \multicolumn{8}{c}{WideResNet34-10} \\ \cline{2-9} \cline{11-18} 
                  & Natural & FGSM & PGD$_{20}$ & PGD$_{100}$ & CW & MIM & AA$_{\rm rand}$ & AA & & Natural & FGSM & PGD$_{20}$ & PGD$_{100}$ & CW & MIM & AA$_{\rm rand}$ & AA \\ \hline
AT$_{\rm PGD}$ & 90.46 & 61.62 & 52.79 & 50.19 & 35.03 & 37.64 & 34.60 & 34.10 & & 92.06 & 64.75 & 55.26 & 53.17 & 36.32 & 38.87 & 36.46 & 35.81 \\
AT$_{\rm HF}$     & \textbf{92.56} & \textbf{67.69} & \textbf{59.12} & \textbf{57.03} & \textbf{42.47} & \textbf{45.50} & \textbf{41.64} & \textbf{40.83} & & \textbf{93.16} & \textbf{68.21} & \textbf{60.46} & \textbf{57.92} & \textbf{43.41} & \textbf{46.79} & \textbf{43.70} & \textbf{42.88} \\ \hline
TRADES          & 88.23 & 63.90 & 58.26 & 56.91 & 36.12 & 41.56 & 36.02 & 35.52 & & 88.90 & 66.49 & 59.27 & 57.23 & 37.72 & 43.18 & 38.21 & 37.31 \\
TRADES$_{\rm HF}$ & \textbf{91.31} & \textbf{68.74} & \textbf{61.13} & \textbf{59.72} & \textbf{41.37} & \textbf{46.94} & \textbf{41.05} & \textbf{40.56} & & \textbf{90.73} & \textbf{69.30} & \textbf{62.23} & \textbf{60.33} & \textbf{43.65} & \textbf{47.34} & \textbf{43.35} & \textbf{41.18} \\ \hline
MART            & 89.13 & 63.19 & 57.46 & 55.40 & 35.84 & 41.67 & 35.62 & 35.15 & & 89.67 & 66.36 & 58.64 & 57.51 & 36.44 & 43.52 & 37.96 & 37.12 \\
MART$_{\rm HF}$   & \textbf{90.71} & \textbf{67.64} & \textbf{60.21} & \textbf{58.37} & \textbf{40.18} & \textbf{47.07} & \textbf{40.93} & \textbf{40.10} & & \textbf{91.02} & \textbf{69.45} & \textbf{62.41} & \textbf{60.62} & \textbf{42.36} & \textbf{48.27} & \textbf{41.82} & \textbf{41.29} \\ \hline
AWP             & 89.64 & 64.05 & 59.03 & 57.76 & 36.51 & 42.45 & 37.74 & 36.89 & & 90.24 & 65.50 & 60.23 & 58.21 & 37.83 & 43.72 & 39.62 & 38.86 \\
AWP$_{\rm HF}$    & \textbf{91.69} & \textbf{68.63} & \textbf{61.68} & \textbf{60.89} & \textbf{41.58} & \textbf{47.82} & \textbf{42.04} & \textbf{41.24} & & \textbf{91.56} & \textbf{69.58} & \textbf{62.85} & \textbf{61.52} & \textbf{42.72} & \textbf{49.36} & \textbf{44.22} & \textbf{43.03} \\ \hline
HELP            & 90.34 & 62.65 & 56.05 & 53.87 & 36.23 & 40.67 & 34.97 & 34.64 & & 91.34 & 67.65 & 57.32 & 54.93 & 37.26 & 42.15 & 38.71 & 37.89 \\
HELP$_{\rm HF}$   & \textbf{91.79} & \textbf{66.04} & \textbf{59.60} & \textbf{57.77} & \textbf{42.86} & \textbf{45.52} & \textbf{41.78} & \textbf{41.23} & & \textbf{92.79} & \textbf{69.70} & \textbf{61.64} & \textbf{58.64} & \textbf{43.38} & \textbf{46.85} & \textbf{43.34} & \textbf{41.83} \\ \hline
\end{tabular}}
\label{svhn-bl}
\end{table*}

Based on the Tab. \ref{cifar100-bl} and Tab. \ref{svhn-bl}, as well as the relevant result on the CIFAR-10 dataset in the main text, it can be demonstrated that HFAT achieves significant improvements across different datasets, defense methods, and network architectures, thus verifying the generality of the strategy.

\end{document}